\documentclass[10pt,twocolumn,letterpaper]{article}

\usepackage{iccv}

\usepackage[pagebackref=true,breaklinks=true,letterpaper=true,colorlinks,bookmarks=false]{hyperref}

\iccvfinalcopy 

\def\iccvPaperID{****} 

\ificcvfinal\fi

\usepackage{xspace}
\usepackage{color}

\newcommand{\mypartitle}[2][2.5]{\vspace*{-#1 ex}~\\{\noindent {\bf #2}}}

\usepackage{url}
\usepackage{amsmath}
\usepackage{graphicx}
\usepackage{paralist}
\usepackage{ amssymb }
\usepackage{algorithm}
\usepackage[noend]{algpseudocode}
\usepackage{epsfig}

\begin{document}
\graphicspath{{IMAGES/}}

\title{Improving speaker turn embedding by \\crossmodal transfer learning from face embedding}

\author{Nam Le$^{1, 2}$, Jean-Marc Odobez$^{1, 2}$\\
$^1$ Idiap Research Institute, Martigny, Switzerland\\
$^2$ \'{E}cole Polytechnique F\'{e}d\'{e}ral de Lausanne, Switzerland\\
{\tt\small\{nle, odobez\}@idiap.ch}
}

\maketitle

\begin{abstract}
Learning speaker turn embeddings has shown considerable improvement in situations where conventional speaker modeling approaches fail.
However, this improvement is relatively limited when compared to the gain observed in face embedding learning, 
which has been proven very successful for face verification and clustering tasks.
Assuming that face and voices from the same identities share some latent properties (like age, gender, ethnicity), 
we propose three transfer learning approaches to leverage the knowledge from the face domain (learned from thousands of 
images and identities) for tasks in the speaker domain.
These approaches,  namely target embedding transfer, relative distance transfer, and clustering structure transfer, 
utilize the structure of the source face embedding space at different granularities to regularize the target 
speaker turn embedding space as optimizing terms. 
Our methods are evaluated on two public broadcast corpora and yield promising advances over competitive baselines 
in verification and audio clustering tasks, especially when dealing with short speaker utterances.
The analysis of the results also gives insight into characteristics of the embedding spaces and shows their potential applications.
\end{abstract}


\section{Introduction}

As the daily production of  broadcast TV and internet content is growing quickly everyday, it is an essential task to make large multimedia corpora easily accessible through search and indexing.
Therefore, research effort has been devoted to unsupervised segmentation of videos into homogeneous segments according to person identity, one of which is speaker diarization, \textit{i.e.} segmenting an audio stream according to the identity of the speaker.
It allows search engines to answer the question ''who speaks when?'' and to create rich transcription of ''who speaks what?'', which is very useful for multimedia documents structuring and indexing.

In the literature, state-of-the-art Gaussian-based speaker diarization methods have been shown to be successful in various types of content such as radio or TV broadcast news, telephone  conversation and meetings \cite{ma2007finding,jousse2009automatic,poignant2014unsupervised}.
In these contents, the speech signal is mostly prepared speech and clean audio, the number of speakers is limited, and the duration of speaker turn is more than 2 seconds on average.
When these conditions are not valid, in particular the assumption of speaker turn duration, the quality of speaker diarization deteriorates~\cite{sarkar2012study}.
As shown in TV series or movies, state-of-the-art approaches do not perform well~\cite{clement2011speaker,bost2014constrained} when there are many speakers (from 28 to 48 speakers), or speaker turns are spontaneous and short (1.6 seconds on average in the Game of Thrones TV series).
To alleviate these shortcomings of speaker diarization, researches have been proposed along two fronts: better methods to learn speaker turn embeddings or utilizing the multimodal nature of video content. The recent work on speaker turn embedding using triplet loss shows certain improvements~\cite{Bredin2017}. Other multimodal related works focus on late fusion of two streams by propagating labels~\cite{bendris2014multiple,bredin2016improving} or high level information such as distances or overlapping duration~\cite{Gay:ICASSP:2014,sargent2016puc}.

In this work, we unite the two fronts by proposing crossmodal transfer learning from a face embedding to improve a speaker turn embedding.
Indeed recently, learning face embeddings has made significant achievements in all tasks, including recognition, verification, and clustering \cite{Schroff2015,parkhi15deep}.
To transpose these advances to the speaker diarization domain, a neural network for speaker turn embedding trained with triplet loss (\textit{TristouNet}) was proposed in~\cite{Bredin2017}. 
Nevertheless, the improvement of this network architecture over the Gaussian-based methods was quite incremental compared
to the gain obtained when using such methods in learning face embeddings.
%
To explain this disparity between modalities, one can point to the clear difference in amounts of training data, 
as there are hundreds of thousands images from thousands identities in any standard face dataset.
The limited size of speech data is very challenging to overcome because we cannot use Internet search engines to collect speech segments similarly to face images in~\cite{parkhi15deep,yi2014learning}. Moreover, manual labeling speech segments is much more costly.
To mitigate the need of massive dataset, we take advantage of pretrained face embeddings by relying on the multimodal nature of person diarization.

Although transfer learning is widely applied in other topics~\cite{zhuang2011exploiting,long2010transfer}, transferring between acoustic and visual domains has mainly been applied to the task of speech recognition~\cite{Moon+15a}, in which the two streams are highly correlated. 
On the other hand, with respect to identity, because there is not a definite one-one inference from a face to a voice, it is still an open question of how to apply transfer learning between a face embedding and a speaker embedding.
To answer this question, we start with an observation. Although one cannot find the exact voice of a person given only a face, however, if given a small set of potential candidates, it is possible to pick a voice which is more likely to come from the given face than other voices.
For example, when most candidates are male voices then it is more likely to find the correct one if the voice is female.
Thus, there are latent attributes which are shared between the two modalities. 
Rather than relying on multimodal data with explicit shared labels such as genders, ages, or accent and ethnicity, we want to discover the latent commonalities from the source domain, a face embedding, and transfer to the target domain, a speaker turn embedding.
Therefore, we hypothesize that by enforcing the speaker turn embedding to have the same geometric properties with  
the face embedding with respect to identity, we can improve the performance of the speaker turn embedding.

Because from one space, there are different properties to be used as constrains to be enforced on the other space, we propose 3 different strategies aiming at different granularity for transferring:
\begin{compactitem}
\item Target embedding transfer: We are given the identity correspondences between the 2 modalities. Hence, given the 2 inputs from the same identity, one can force the desired embedded features of the speaker turn to be close to embedded features of the face. Minimizing the disparity between the 2 embedding spaces with respect to identity will act as a regularizing term for optimizing the speaker turn embedding.

\item Relative distance transfer: One can argue that exact similar location in the embedding spaces is hard to achieve given the fuzzy relationship between the 2 modalities. It may be sufficient to only enforce relative order between identities, thus assuming that 2 people who look more similar will have more similar voices.

\item Clustering structure transfer: This approach focus on discovering shared commonalities between the 2 embedding spaces such as age, gender, or ethnicity. If a group of people share common facial traits, we expect their voices to also share common acoustic features. In particular, the shared common traits in our case is expressed as belonging to the same cluster of identities in the face embedding space.
\end{compactitem}
Experiments conducted on 2 public datasets REPERE and ETAPE show significant improvement over the competitive baselines, especially when dealing with short utterances.
Our contributions are also supported by crossmodal retrieval experiments and the visualization of our intuition.

The rest of the paper is organized as follows. Sec.~\ref{sec:related_work} reviews other works related to ours, Sec.~\ref{sec:triplet} introduces triplet loss and the motivation of our work, Sec.~\ref{sec:methods} describes our transfer methods in details.
Sec.~\ref{sec:experiments} presents and discusses the experimental results, while Sec.~\ref{sec:conclusion} concludes the paper.

\section{Related Work}
\label{sec:related_work}

Below we discuss prior works on audio-visual person recognition and transfer learning which share similarities with our proposed methods.

As person analysis tasks in multimedia content such as diarization or recognition are multimodal by nature, significant effort has been devoted to using one modality to improve another.
Several works exploit labels from the modality that has superior performance to correct the other modality. In TV news, as detecting speaker changes 
produces less false alarm rate and less noise than detecting and clustering faces, speaker diarization hypothesis is used to constrain face clustering,
\textit{i.e.} talking faces with different voice labels should not have the same name~\cite{bendris2014multiple}.
Meanwhile in ~\cite{bredin2016improving}, because face clustering outperforms speaker diarization in TV series, labels of face clusters are propagated to the corresponding speaker turns.
Another approach is to perform clustering jointly in the audio-visual domain.~\cite{sargent2016puc} linearly combines the acoustic distance and the face representation distance of speaking tracks to perform graph-based optimization; while ~\cite{Gay:ICASSP:2014} formulates the joint clustering problem in a CRF framework with the acoustic distance and the face representation distance as pair-wise potential functions.
Beside late fusion of labels, early fusion of features proposed in~\cite{Hu2015a, Ren2016} is only suitable for supervised tasks; and because their datasets are limited with 6 identities, the case is not conclusive.
Note that the aforementioned works focus on aggregating two streams of information whereas we emphasize on the transfer of knowledge from one embedding space to another.
By applying recent advances in embedding learning, with deep networks for face \cite{parkhi15deep, Schroff2015} and speaker turn \cite{Bredin2017}
our goal is not only to improve the target task (as speaker turn embedding in our case) but also provide a unified way for multimodal combination.

Each of our three learning approaches draw inspiration from a different line of research. 
First, we can point to coupled matching of image-text or heterogeneous recognition \cite{li2011cross,hu2016multimodal,liong2016deep} or harmonic embedding~\cite{Schroff2015} as related background for our target embedding transfer. Since it is arguable that audio-visual identities contain less correlated information, our method uses the one-one correspondence as a regularization term rather than as an optimal goal.
Second, as the learning targets is an Euclidean embedding space in both modalities, relative distance transfer is inspired by metric imitation~\cite{dai2015metric} or multi-tasks metric learning~\cite{bhattarai2016cp}. In our work, the triangular relationship is transferred across modalities instead of neighbourhood structure or across tasks of the same modality.
Finally, though co-clustering information and cluster correspondence inference have been used in transfer learning on traditional tasks of text mining~\cite{zhuang2011exploiting,long2010transfer}, we are first to expand that concept into exploiting clustering structure of person identities for crossmodal learning.

\section{Triplet loss and motivation}
\label{sec:triplet}
Given a labeled training set of $\{(x_i, y_i)\}$, in which $x_i \in \mathbb{R}^D, y_i \in \{1, 2, .., K\}$, we define an embedding as  $f(x) \in \mathbb{R}^d$, which maps an instance $x$ into a $d$-dimensional Euclidean space. Additionally, this embedding is constrained to live on the $d$-dimensional hypersphere, \textit{i.e.} $||f(x)||_2 = 1$. Within the hypersphere, the distance between 2 projected instances is simply the Euclidean distance:
\begin{equation}
d(f(x_i), f(x_j)) = ||f(x_i) - f(x_j) ||_2
\end{equation}

In this new embedding space, we want the intra-class distances $d(f(x_i), f(x_j)), \forall x_i, x_j / y_i = y_j$ to be minimized and the inter-class distances $d(f(x_i), f(x_j)), \forall x_i, x_j / y_i \neq y_j$ to be maximized.
A major advantage of embedding learning is that the projection $f$ is class independent. At test time, we can expect examples from a different class, or identity, to appear and still satisfy the embedding goals. This makes embedding learning suitable for verification and clustering tasks.

\begin{figure}[!htb]
 \centering
 \includegraphics[height=30mm]{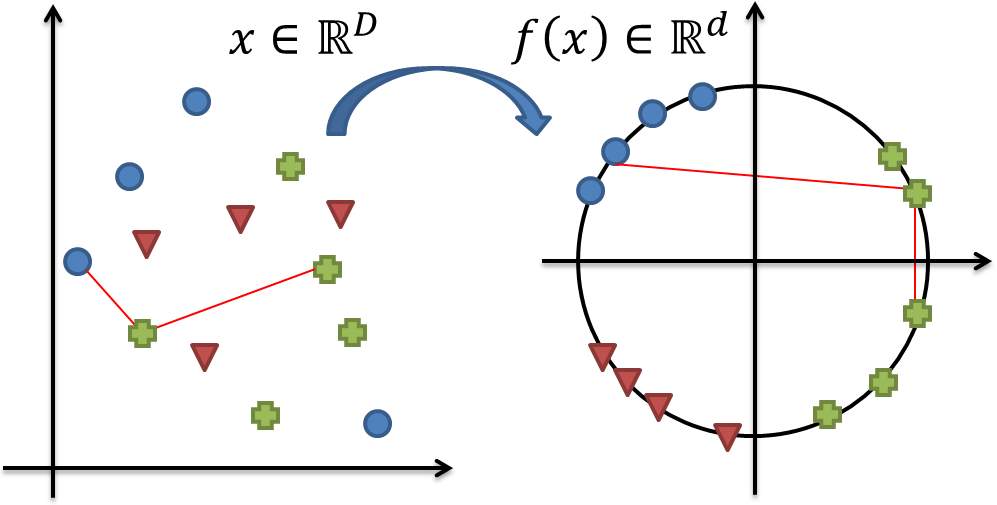}

 \caption{Illustration of an embedding space.}
 \label{fig:triplet-example}
\end{figure}

To achieve such embedding, one method is to learn the projection that optimizes the triplet loss in the embedding space. 
A triplet consists of 3 data points: $(x_a, x_p, x_n)$ such that $y_a = y_p$ and $y_a \neq y_n$ and thus, we would like the 2 points $(x_a, x_p)$ to be close together and the 2 points $(x_a, x_n)$ to be further away by a margin $\alpha$ in the embedding space
\footnote{The value of $\alpha$ varies depending on the particular loss function to optimize (such as $\mathcal{L^A}, \mathcal{L^V}$, or $\mathcal{L}^{V\rightarrow A}$). In this paper we use one value of $\alpha=0.2$ in all cases.}. Formally, a triplet must satisfy:
\begin{equation}
d(f(x_a), f(x_p)) + \alpha < d(f(x_a), f(x_n)),
\forall (x_a, x_p, x_n) \in T
\end{equation}
where $T$ is the set of all possible triplets of the training set, and $\alpha$ is the margin enforced between the positive and negative pairs.
Subsequently, we define the loss to be minimized as:
\begin{equation}
\label{eq:loss_emb}
\mathcal{L}(f) = \frac{1}{|T|}\sum_{(x_a, x_p, x_n) \in T} {l(x_a, x_p, x_n, f)}
\end{equation}

in which 
\begin{multline}
\label{eq:tripletloss}
l(x_a, x_p, x_n; f) = \\ \max\{0, d(f(x_a), f(x_p)) - d(f(x_a), f(x_n)) + \alpha\}
\end{multline}
Fig. \ref{fig:triplet-example} shows an example of an embedding space, in which samples from difference classes are separated. By choosing $d << D$, one can learn a projection to a space that is both distinctive and compact.

In spite of its advantages, the triplet loss training is empirical and depends on the training data, the initialization, and triplet sampling methods.
For a certain set of training samples, there can be an exponential number of possible solutions that yield the same training loss.
One approach to guarantee good performance is to make sure that the training data come from the same distribution of the test data (as in \cite{parkhi15deep}). 
Another solution for the projection to work in more general unseen cases may be to gather a massive training dataset with more training data (which is the case of FaceNet which was trained with 100-200 millions images of 8 millions of identities~\cite{Schroff2015}). 
Although it is possible to gather such a large scale dataset for visual information, it is less the case for acoustic data. This explains why speaker turn embedding \textit{TristouNet} only gains slight improvement over Gaussian-based methods~\cite{Bredin2017}. To alleviate the data concern, we tackle the problem of embedding learning from the multimodal point of view. By using a superior face embedding network that was trained on a face dataset with the same identities as in the acoustic dataset, we can regularize the speaker embedding space and thus guide the training process to a better minima.

\section{Crossmodal transfer learning}
\label{sec:methods}
In audio-visual (or multimodal data in general) settings, data contain 2 corresponding streams $\{(x^A_i, x^V_i, y_i)\}$. If the learning process is applied independently to each modality, we can learn 2 projections $f_A$ and $f_V$ into 2 embedding spaces $\mathbb{R}^{d_A}$ and $\mathbb{R}^{d_V}$ following their own respective losses:
\begin{equation}
\label{eq:audio_emb}
\mathcal{L}^A(f^A) = \frac{1}{|T^A|}\sum_{(x^A_a, x^A_p, x^A_n) \in T^A}{l(x^A_a, x^A_p, x^A_n; f^A)}
\end{equation}
and
\begin{equation}
\label{eq:face_emb}
\mathcal{L}^V(f^V) = \frac{1}{|T^V|}\sum_{(x^V_a, x^V_p, x^V_n) \in T^V}{l(x^V_a, x^V_p, x^V_n; f^V)}
\end{equation}
in which $\mathcal{L}^A $ and $\mathcal{L}^V$ are defined from the general embedding loss Eq.~\ref{eq:loss_emb} to speaker turn embedding and face embedding.

As shown in the experiments, $f^V$ can already achieve significantly lower than the counterpart in acoustic domain, therefore our goal is to transfer the knowledge from face embedding to the speaker turn embedding. Hence, we assume that $f^V$ is already trained with Eq.~\ref{eq:face_emb} using the corresponding face dataset (as well as optional external data).  
Using $f_V$, an auxiliary term $\mathcal{L}^{V\rightarrow A}(f^A)$ is defined to regularize the relationship between voices and faces from the same identity in addition to the loss function used to train speaker turn embedding in Eq.~\ref{eq:loss_emb}. Formally, the final loss function can be written as:
\begin{equation}
\label{eq:fullloss}
\mathcal{L}(f^A) = \mathcal{L}^A(f^A) + \lambda \mathcal{L}^{V\rightarrow A}(f^A)
\end{equation}
The transfer loss $\mathcal{L}^{V\rightarrow A}(f^A)$ depends on what type of knowledge is transferred across modalities. 
$\lambda$ is a constant hyper-parameter chosen through experiments specifically for each transfer type. 
In the following sections, different types of $\mathcal{L}^{V\rightarrow A}(f^A)$ will be described in details.

\subsection{Target embedding transfer}

Assuming that $f^A$ projects $x_i^A$ into the same hypersphere as $f^V(x_j^V)$, one can observe that by enforcing $f^A(x_i^A)$ to be in close proximity of $f^V(x_j^V)$ when $y_i = y_j$, $f^A$ could achieve a similar training loss as $f^V$.
In that case, the  regularizing term  in Eq.~\ref{eq:fullloss} can be defined 
as the disparity between crossmodal instances of the same identity:
\begin{equation}
\label{eq:coupled}
\mathcal{L}^{V\rightarrow A}(f^A) = \sum_{(x_i^A, x_j^{V}) / y_i = y_j}{d(f^A(x_i^A), f^V(x_j^V))}
\end{equation}
The goal of Eq.~\ref{eq:coupled} is to minimize intra-class distances by binding embedded speaker turns 
and embedded faces within the same class similarly to coupled multimodal projection methods~\cite{li2011cross,liong2016deep}. 
In this work, we extend this goal further by adopting the multimodal triplet paradigm
to jointly minimize intra-class distances and maximize inter-class distances.

{\noindent \bf Multimodal triplet loss.}
In addition to minimizing the audio triplet loss of Eq.~\ref{eq:audio_emb}, we also 
want two embedded instances to be close if they come from the same identity, 
regardless of the modality they comes from, and to be far from embedded instances of all other identities
in both modalities as well.
Concretely, the regularizing term is thus defined as the triplet loss over multimodal triplets:
\begin{multline}
\label{eq:target_emb}
\mathcal{L}^{V\rightarrow A}(f^A) = \\ \frac{1}{|T_{tar}|}\sum_{(x^{m_a}_a, x^{m_p}_p, x^{m_n}_n) \in T_{tar}}{l(x^{m_a}_a, x^{m_p}_p, x^{m_n}_n; f^{A},f^{V})}
\end{multline}
where $m_{\bullet}$ is the modality associated with the sample $x_{\bullet}^{m_{\bullet}}$, and 
the loss $l$ is adapted from Eq.~\ref{eq:tripletloss} by using the embedding appropriate to each sample modality.
The set  $T_{tar}$ denotes all useful and valid cross-modal triplets, i.e. with the positive sample to be of the same identity of the anchor ($y_a = y_p$), and the negative sample to be from another identity ($y_a \neq y_n$); 
and with  $(m_a, m_p, m_n) \in Q_{A,V}$, the set of valid modalities (all combinations except $(V, V, V)$, 
$(V, V, A)$, and $(A, A, A)$ already considered in the primary loss of Eq. \ref{eq:audio_emb}). 
For instance, if $(m_a, m_p, m_n) = (A, V, V)$, the loss 
will foster the decrease of the intra-class distance between $f^A(x^A_a)$ and $f^V(x^V_p)$ 
while increasing  the inter-class distance between $x^A_a$ and $x^V_n$. 
The strategy to collect the set $T_{tar}$ at each epoch of the training  
is described in Alg.~\ref{algo:target}.

\begin{algorithm}[tb]
  \caption{Target embedding transfer triplet set.
    \label{algo:target}}
  \begin{algorithmic}[1]
  		\State{\textbf{Input} $f^A$, $f^V$, $Q_{A,V}$, $\{(x^A_i, x^V_i, y_i)\}_{i=1..N}$ }
      \State{$T_{tar} = \emptyset$}
	  \For{$\forall (a, p ,n) / y_a = y_p \wedge y_a \neq y_n$}
	  		\For{$m_a, m_p, m_n \in \{Q_{A,V}\}$}
	  		    \State{$d_{a,p}=d(f^{m_a}(x^{m_a}_a), f^{m_p}(x^{m_p}_p))$}
	  		    \State{$d_{a,n}=d(f^{m_a}(x^{m_a}_a), f^{m_n}(x^{m_n}_n))$}
				\If{$ d_{a,p} + \alpha > d_{a,n}$}
				    \State{$T_{tar} = T_{tar} \cup (a, p ,n)$}
  				\EndIf
  	  		\EndFor
  	  \EndFor	
  	 \State{\textbf{Output} $T_{tar}$}
  \end{algorithmic}
\end{algorithm}

Using Eq.~\ref{eq:target_emb} as regularizing  term in  $\mathcal{L}(f^A)$, 
one can effectively use the embedded faces as targets to learn a speaker turn embedding. 
Note that this is similar in spirit to the  neural network distillation~\cite{hinton2015distilling}, using one embedding as a teacher for the other.
Moreover, the two modalities can be combined straightforwardly as their embedding spaces can be 
viewed as one harmonic space~\cite{Schroff2015}.

\subsection{Relative distance transfer}
\label{subsec:relative}

The correspondence between faces and voices is not a definitive one-to-one, \textit{i.e.} 
it is not trivial to precisely select the face corresponding to a voice one  has heard.
Therefore target embedding transfer might not generalize well even when achieving low training error.
Instead of the exact locations, the relative distance transfer approach works at a 
lower granularity and aims to mimic the discriminative power (i.e. the notion of being close or far) 
of the face embedding space.
Thus, it does not directly transfer the embeddings individual instances but the relative distances between their identities.

Before computing relative distances, let us define the mean face representation $M_{y}$ of a person and  
the distance between identities within the face embedding space according to:
\begin{equation} 
M_{y} = \frac{1}{|X_y|}\sum_{x_i \in X_y} f^V(x_i)
\mbox{ and }
d(y_i, y_j) = d(M_{y_i}, M_{y_j}), 
\end{equation}
where $X_y$ is the set of visual samples with identity $y$. 
The goal is then to collect in the set $T_{rel}$ all audio triplets  $(a, p, n)$ {\em with arbitrary identities} 
where the sample $p$ has an identity which is closer to the identity of the anchor sample $a$ than 
the identity of the sample $n$,  as defined in the face embedding.
In other words, if within the face embedding space the relative distances among the 3 
identities of the triplet $(a, p, n)$ follows:
\begin{equation}
\label{eq:source_cond}
d(M^V_{y_a}, M^V_{y_p}) < d(M^V_{y_a}, M^V_{y_n}), 
\end{equation}
then this relative condition must hold in the speaker turn embedding space as well:
\begin{equation}
\label{eq:target_cond}
d(f^A(x^A_a), f^A(x^A_p)) + \alpha < d(f^A(x^A_a), f^A(x^A_n))
\end{equation}
Then, at each epoch, Eq.~\ref{eq:source_cond} and ~\ref{eq:target_cond} can be used 
to collect the set $T_{rel}$, as shown in Alg.~\ref{algo:relative}, and the regularizing transfer 
loss $\mathcal{L}^{V\rightarrow A}(f^A)$ can then be defined as the average sum of the standard triplet loss over this set. 
In theory, relative distance transfer can achieve the same training error as with target embedding transfer,
but leave more freedom to the relaxation of the exact location of the embedded features.

\begin{algorithm}
  \caption{Relative distance transfer triplet set.
    \label{algo:relative}}
  \begin{algorithmic}[1]
  	  \State{\textbf{Input} $f^A$, $f^V$, $\{M^{y}\}_{y=1..K}$, $\{(x^A_i, x^V_i, y_i)\}_{i=1..N}$ }
      \State{$T_{rel} = \emptyset$}
	  \For{$\forall (a, p ,n) / y_a \neq y_p \wedge y_a \neq y_n$}
     	  \If{$d(M^V_{y_a}, M^V_{y_p}) < d(M^V_{y_a}, M^V_{y_n})$}
				\State{$d_{a,p}=d(f^A(x^A_a), f^A(x^A_p))$}
	  		    \State{$d_{a,n}=d(f^A(x^A_a), f^A(x^A_n))$}
	  		    \If{$d_{a,p} + \alpha > d_{a,n}$}
				    \State{$T_{rel} = T_{rel} \cup (a, p ,n)$}
  				\EndIf
  	  		\EndIf
  	  \EndFor	
  	  \State{\textbf{Output} $T_{rel}$}	
  \end{algorithmic}
\end{algorithm}

\subsection{Clustering structure transfer}

The common idea of the 2 previous transferring methods is that people with similar faces should have similar voices. 
Thus they aim at putting constrains based on the distances among individual instances in the face embedding space. 
In clustering structure transfer, the central idea does not focus on pair of identities.
but rather, we hypothesize that commonalities between 2 modalities can be discovered amongst groups of identities. 
For example, people within a similar age group are more likely to be close together in the face embedding space, 
and we also expect them to have more similar voices in comparison to other groups.

Based on this hypothesis, we propose to regularize the target speaker turn embedding space 
to have the same clustering structure with the source face embedding space.
To achieve that, we first discover groups in the face embedding space by  performing a K-Means clustering on the set of mean identity representations $\{M^V_{y_i}\}$ obtained as in Sec.~\ref{subsec:relative}. 
If we denote by $C$ the number of clusters, the resulting cluster mapping function is defined as: 
\begin{align*}
g_m: \{1..K\} &\rightarrow \{1..C\} \\
y &\rightarrow c_y
\end{align*}
Secondly, to define  the regularizing term $\mathcal{L}^{V\rightarrow A}(f^A)$, 
we simply consider the set of cluster labels $c_{y_i}$ attached to each audio sample  $(x^A_i, y_i)$
 as the second label, and define accordingly a triplet loss relying on this second 
label (i.e by considering the instances $(x^A_i,c_{y_i})$).
In this way, one can guide the acoustic instances of identities from the same cluster to be close together, 
thus preserving the source clustering structure. How to collect the set of triplet $T_{str}$ to be 
used for the regularizing term at each epoch is detailed in Alg.\ref{algo:structure}.

\begin{algorithm}
  \caption{Clustering struct. transfer triplet set.
    \label{algo:structure}}
  \begin{algorithmic}[1]
      \State{\textbf{Input} $f^A$, $f^V$, $g_m$, $\{(x^A_i, x^V_i, y_i)\}_{i=1..N}$ }
      \State{Cluster mapping $g_m$: $y \rightarrow c_{y}, \forall y \in {1 \ldots K}$}
      \State{$T_{str} = \emptyset$}
	  \For{$\forall (a, p ,n) / c_{y_a} \neq c_{y_p} \wedge c_{y_a} \neq c_{y_n}$}
				\State{$d_{a,p}=d(f^A(x^A_a), f^A(x^A_p))$}
	  		    \State{$d_{a,n}=d(f^A(x^A_a), f^A(x^A_n))$}
	  		    \If{$d_{a,p} + \alpha > d_{a,n}$}
				    \State{$T_{str} = T_{str} \cup (a, p ,n)$}
  				\EndIf
  	  \EndFor
  	  \State{\textbf{Output} $T_{str}$}	
  \end{algorithmic}
\end{algorithm}

This group structure can be expected to generalize for new identities because even though a person is unknown, 
he/she belongs to a certain group which share similarities in the face and voice domains.
In our work, we only apply K-Means once on the mean facial representations. 
However, as people usually belong to multiple non-exclusive common groups, each with  a different attribute, 
it would be interesting in further works to aggregate multiple clustering partitions with different 
initial seeds or with different number of clusters.
As the space can be hierarchically structured, one other possibility could be to apply 
hierarchical clustering to obtain these multiple partitions.

\section{Experiments}
\label{sec:experiments}
We first describe the datasets and evaluation protocols before discussing the implementation details and the results.
For the reproducibility, our annotations, pretrained models, and auxiliary scripts will be made publicly available.

\subsection{Datasets}

\mypartitle{REPERE~\cite{giraudel2012repere}.} We use this standard dataset to collect people tracks with corresponding voice-face information. It features programs including news, debates, and talk shows from two French TV channels, LCP and BFMTV, along with annotations available through the REPERE challenge. 
The annotations consist of the timestamps when a person appears and talks. By intersecting the talking and appearing information, we can obtain all segments with face and voice from the same identity. As REPERE only contains sparse reference bounding box annotation, automatic face tracks 
are aligned with reference bounding boxes to get the full face tracks.
This collection process is followed by manual examination for correctness and consistency and to remove short tracks (less than 18 frames $\approx$ 0.72s).
The resulting data is split into training and test sets. 
Statistics are shown in Tab.~\ref{tab:repere_stats}.

\begin{table}[tb]
\centering
\vspace*{-2mm}
\caption{Statistics of tracks extracted from REPERE. The training and test sets have disjoint identities.}
\begin{tabular}{c|c|c|c|}
\cline{2-4}
																& \# shows	& \# people	& \# tracks \\ \hline
 \multicolumn{1}{|c|}{training} & 98				& 208 			& 1876 			\\ \hline
 \multicolumn{1}{|c|}{test}			& 35				& 98				& 629  			\\ \hline
								
\end{tabular}
\vspace*{-3mm}
\label{tab:repere_stats}
\end{table}

\mypartitle{ETAPE~\cite{gravier2012etape}.} This standard dataset contains 29 hours of TV broadcast. 
In this paper, we only consider the development set to compare with state-of-the-art methods. 
Specifically, we use similar settings for the ''same/different'' audio experiments than in~\cite{Bredin2017}. 
From this development set, 5130 1-second segments of 58 identities are extracted. Because 15 identities appear in the REPERE training set, we remove them and retain 3746 segments of 43 identities.

\subsection{Experimental protocols and metrics}

\mypartitle{Same/different experiments.} 
Given a set of segments, distances between all pairs are computed. 
One can then decide if a pair of instances has the same identity if their (embedded) distance is below a threshold. 
We can then report the equal error rate (EER), \textit{i.e.} the value when the 
false negative rate and the false positive rate become equal as we vary the threshold.

\mypartitle{Clustering experiments.} 
From a set of all audio (or video) segments, a standard hierarchical clustering is applied using the distance between cluster means in the embedded space as merging criteria. 
Then, each time 2 clusters are merged, we compute 3 metrics on the clustering set: 
\begin{compactitem}
\item Weighted cluster purity (WCP) \cite{tapaswi2014total}: For a given set of clusters $C = \{c\}$, each cluster $c$ has a weight of $n_c$, which is the number of segments within that cluster. 
At initialization, we start from $N$ segments with weight $1$ each. 
The purity $purity_c$ of a cluster $c$ is the fraction of the largest number of segments from the same 
identities to the total number of segments in the cluster $n_c$. WCP is calculated as:
\[
WCP = \frac{1}{N}\sum_{c \in C} n_c \cdot purity_c 
\]

\item Weighted cluster entropy (WCE): A drawback from WCP is that it does not distinguish the errors.
For instance, a cluster with 80\% purity, 20\% error due to 5 different identities is more severe 
than if it is only due to 2 identities. 
To characterize this point, we thus compute the entropy of a cluster, from which WCE is calculated as:
\[
WCE = \frac{1}{N}\sum_{c \in C} n_c \cdot entropy_c
\]

\item Operator clicks index (OCI-k)~\cite{guillaumin2009you}: This is the total number of clicks required to label all clusters. If a cluster is 100\% pure, only 1 click is required. Otherwise, besides 1 click to annotate segments of the dominant class, then 1 extra click is needed to correct each erroneous track of a different class. 
For a cluster $c$ of $n_c$ speaker segments, the cluster cost is  formally defined as:
\[
\mbox{OCI-k}(c) = 1 + (n_c - max(\{n^c_i\})),
\]
where $n^c_i$ denotes the number of segments from identity $i$ in the cluster. 
The cluster clicks are then added to produce the overall OCI-k performance measure.
This metric simultaneously combines the number of clusters and cluster quality in one number to represent the  
manual effort in practical applications.
\end{compactitem}

\subsection{Implementation details}

\mypartitle{Face embedding.} Our face model is based on ResNet-34~\cite{he2016deep} trained on CASIA-WebFaces~\cite{yi2014learning}. We follow the procedure of~\cite{parkhi15deep} as follows:
%
\begin{compactitem}
\item A DPM face detector~\cite{dubout2013deformable} is run to extract a tight bounding box around each face. No further preprocessing is performed except for randomly flipping training images.

\item ResNet-34 is first trained to predict 10,575 identities by minimizing cross entropy criteria. Then the last layer is removed and the weights are frozen. 

\item The last embedding layer with a dimension of $d=128$ is learned using Eq.~\ref{eq:face_emb} and the face tracks of the  REPERE training set. 
\end{compactitem} 

\mypartitle{Speaker turn embedding.} 
Our implementation of \textit{TristouNet} consists of a bidirectional LSTM with the hidden size of 32. 
It is followed by an average pooling of the hidden state over the different time steps of the audio sequence, 
followed by 2 fully connected layers of size 64 and 128 respectively. 
As input acoustic features to the LSTM, 13 Mel-Frequency Cepstral Coefficients (MFCC) are extracted with energy and their first and second derivatives.

\mypartitle{Optimization.} All embedding networks are trained using a fixed $\alpha=0.2$ 
and the RMSProp optimizer~\cite{tieleman2012lecture} with a $10^{-3}$ learning rate. From each mini-batch, both hard and soft negative triplets are used for learning.

\mypartitle{Baselines.} We compare our speaker turn embedding with 3 approaches: Bayesian Information Criterion (BIC)~\cite{chen1998speaker}, Gaussian divergence (Div.)~\cite{barras2006multistage}, and the original \textit{TristouNet}~\cite{Bredin2017}.

\subsection{Experimental results}

\subsubsection{Face embedding}

We conducted this experiment to choose the best (more accurate) face embedding to transfer to the audio domain 
amongst the following candidates:
\begin{compactitem}
\item VGG-Face(dim=4096): We use the model from \cite{parkhi15deep}, which was pretrained using 2.6 millions faces of 2622 identities.
\item Rn34-FC(dim=512): ResNet-34 trained using the CASIA-WebFaces and using the activation of the last layer before the softmax
identity classification as face features.
\item Rn34-Emb(dim=128): The embedding layer is learned using the trained face tracks of the REPERE dataset.
\end{compactitem}

From the REPERE test set, 6000 pairs of tracks (3000 negative, 3000 positive) are selected for benchmarking the embeddings 
using the same/different experimental setting. 
We compare using the EER and the AUC of the ROC curve. 
From Tab.~\ref{tab:face_repere}, we can see that the RestNet34 slightly outperforms VGG-Face, 
and that further using a triplet loss learned using the face tracks of the REPERE data  
helps improving the results. 
Thus in the following experiments, Rn34-Emb is chosen as embedding to transfer to the audio domain.

\begin{table}[tb]
\centering
\caption{Results of face representations on 6000 pairs of REPERE test tracks.}
\begin{tabular}{c|c|c|c|}
\cline{2-4}
																& VGG-Face 	& Rn34-FC & Rn34-Emb 	\\ \hline
\multicolumn{1}{|c|}{AUC - ROC} & 99.02 		& 99.15 	& 99.43 		\\ \hline
\multicolumn{1}{|c|}{EER} 			& 4.35 			& 3.6 		& 3.15 			\\ \hline
\end{tabular}
%
\label{tab:face_repere}
\end{table}

\subsubsection{REPERE - Clustering experiment}

\begin{figure}[tb]
\centering
a)
\epsfig{file=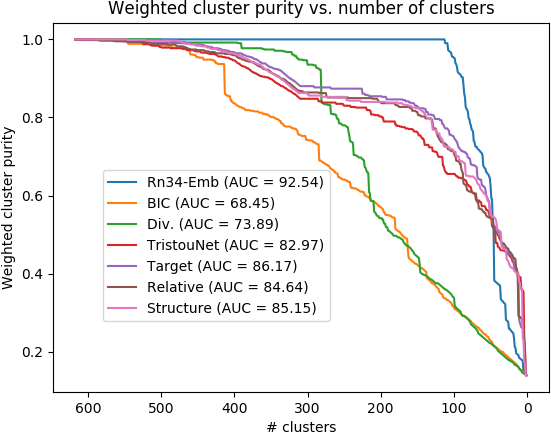,height=50mm}

b)
\epsfig{file=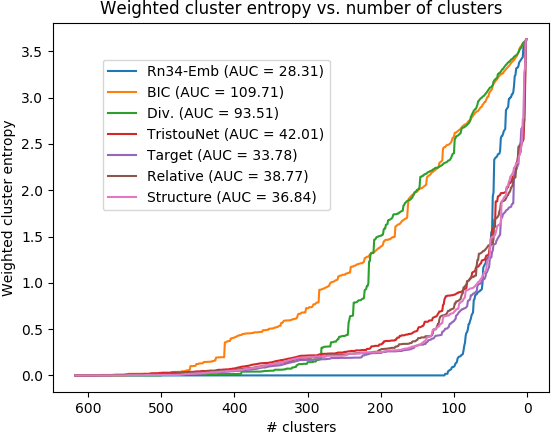,height=50mm}
\caption{Evaluation of hierarchical clustering on REPERE. (a) weighted cluster purity. (b) weighted cluster entropy.}
\label{fig:clustering_graph}
\end{figure}

We applied the audio (or video) hierarchical clustering to the 629 audio-visual test tracks of REPERE.
Results are presented in Fig.~\ref{fig:clustering_graph}. 
Face clustering with Rn34-Emb clearly outperforms all speaker turn based methods. 
At the beginning, Div. first merges longer audio segments with enough data so it achieves higher purity. 
However, as small segments get progressively merged, the performance of BIC and Div. quickly deteriorate due to the lack of good voice statistics.

Our transferring methods surpass \textit{TristouNet} in both metrics, especially in the middle stages, when the distances between clusters becomes more confusing. 
This shows that the knowledge from the face embedding helps distinguishing confusing pairs of clusters. 
The gap in WCE also means that our embedding is also more consistent with respect to the inter-cluster distances. 
We should note that in WCP and WCE, segments count as one unit and are not weighted according to their duration as done in traditional diarization metrics. 
This is one reason while traditional approaches BIC and Div methods  appear much worse with the clustering metrics. 
More experiments on full diarization are needed in future works.

Tab.~\ref{tab:ocik-repere} reports the number of clicks to label and correct the clustering results. 
Our target embedding transfer reduces the OCI-k by 30 from the closest competitor in both the best case 
and with the ideal number of clusters.
This in practice can decrease the effort of human annotation by $10-12\%$.
Other transferring methods also show improvement of 7-10\%.

\begin{table}[tb]
\centering
\caption{Result of OCI-k metric on the REPERE test set. 'Min' reports minimum value of OCI-k and its number of clusters. 'At ideal clusters' reports OCI-k at 98 clusters corresponding to 98 identities.}
\begin{tabular}{c|c|c|}
\cline{2-3}																						& Min (\# clusters) & At 98 clusters \\ \hline
 \multicolumn{1}{|c|}{Rn34-Emb (V)} 												& 113 (113)					& 136	\\ \hline
 \multicolumn{1}{|c|}{BIC \cite{chen1998speaker}} 			& 451 (390)					& 525 \\ \hline
 \multicolumn{1}{|c|}{Div. \cite{barras2006multistage}} & 330 (289) 				& 521 \\ \hline
 \multicolumn{1}{|c|}{\textit{TristouNet}~\cite{Bredin2017}} 		& 275 (124) 				& 285 \\ \hline
 \multicolumn{1}{|c|}{Target} 													& 241 (123)					& 255	\\ \hline
 \multicolumn{1}{|c|}{Relative} 												& 256 (132) 				& 268	\\ \hline
 \multicolumn{1}{|c|}{Structure} 												& 255 (132)					& 271	\\ \hline
\end{tabular}
%
\label{tab:ocik-repere}
\end{table}

\subsubsection{ETAPE - Same/different experiment}

From the ETAPE development set, 3746 segments of 43 identities are extracted. 
From these segments, all possible pairs are used for testing  and the EER is reported in Tab.\ref{tab:eer_etape}. 
All of our networks with transferred knowledge outperform the baselines. 
With short segments of 1 second, BIC and Div. do not have enough data to fit the Gaussian models well, therefore they perform poorly. 
By transferring from visual embedding, we can improve \textit{TristouNet} with a relative improvement of 6\% of EER. 
We should remark that in~\cite{Bredin2017}, the original \textit{TristouNet} achieved 17.3\% and 14.4\% when being trained and tested on 1s sequences and 2s sequences respectively. 
However, it is important to note that our models are trained on a smaller dataset (4.5h vs. 13.8h of ETAPE data in \cite{Bredin2017}) and from an independent training set (REPERE vs. ETAPE). Using our transfer learning methods, the speaker turn embedding model could be easily trained by combining different dataset, \textit{i.e.} combining REPERE and ETAPE training sets.

\mypartitle{Comparison of transfer methods.}
Though the difference is small, target embedding shows an advantage in both the REPERE clustering experiments 
and in the ETAPE experiment. It seems that as the level of granularity decreases, the performance decreases. 
It could be interesting in future work to combine these different transfer method to see whether any further 
gain could be obtained.

\begin{table}[tb]
\centering
\caption{EER reported on ETAPE dev set. Note that our V $\rightarrow$ A transfer methods are trained on 1s. sequences (\footnotesize{$^*$ denotes reported results from \cite{Bredin2017}})}
\begin{tabular}{|c|c|c|c|c|c|c|c|}
\cline{1-8}
\multicolumn{2}{|c|}{BIC\cite{chen1998speaker}} & \multicolumn{2}{|c|}{Div.\cite{barras2006multistage}} & \cite{Bredin2017} & \multicolumn{3}{|c|}{V $\rightarrow$ A transfer} \\ \hline
1s. 	& 2s.$^*$ & 1s. 	& 2s.$^*$ & 1s. 	& Tar. & Rel. & Str. \\ \hline
32.4 	& 20.5 		& 28.9 	& 22.5 		& 19.1 	& 18.0 & 18.2 & 18.3 \\ \hline
\end{tabular}
%
\label{tab:eer_etape}
\end{table}

\subsubsection{Parameter sensitivity}

\begin{figure*}[tb]
\centering
a)
\epsfig{file=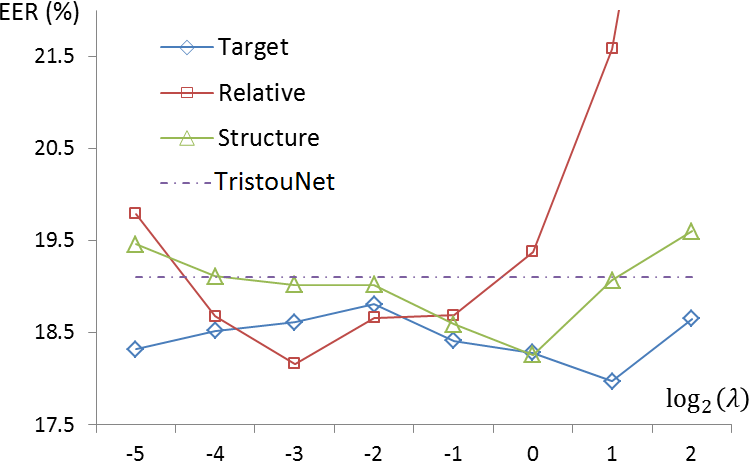,height=33mm}
b)
\epsfig{file=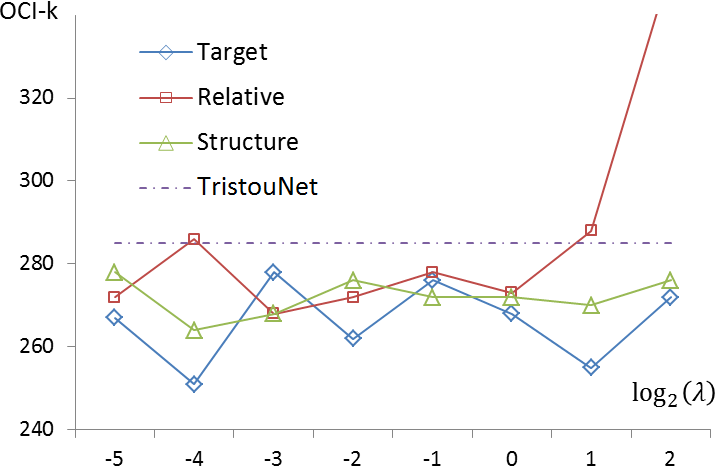,height=33mm}
c)
\epsfig{file=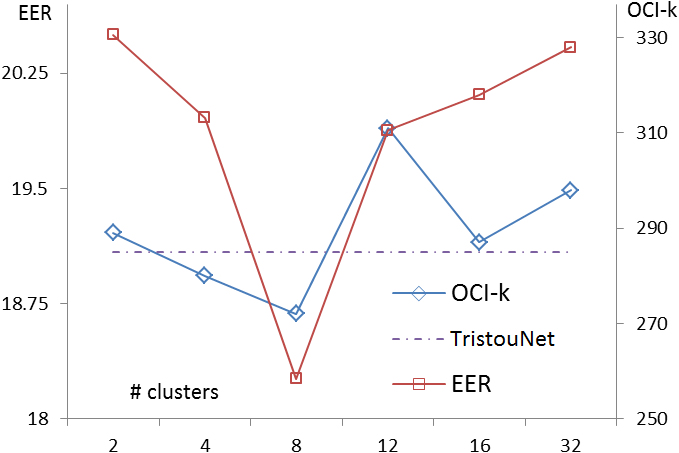,height=33mm}
\caption{Result of different values of hyperparameters. (a)EER on ETAPE as $\lambda$ changes, (b) OCI-k on REPERE as $\lambda$ changes, (c) EER on ETAPE and OCI-k on REPERE as the number of clusters for structure transfer changes.}
\label{fig:beta_graph}
\end{figure*}

In all our transfer learning settings, we need to choose one hyper parameter $\lambda$, 
and the number of clusters for structure transfer setting.
Hence, we perform benchmarking with different values of $\lambda$ 
and report results in Fig.~\ref{fig:beta_graph}. In Fig.~\ref{fig:beta_graph}-(a) and (b), we can observe that except for relative distance transfer, the rest are quite insensitive to this hyper parameter $\lambda$. 
Each of them has a different optimal value, which is due to the  difference in the nature of each method. 
One possible explanation for the case of relative distance transfer when $\lambda \geq 2$ is that there is no proximity constrains on the location of the embedded features, thus instability is not bounded and can increase at test time.
Fig.\ref{fig:beta_graph}-(c) shows how structure transfer performs under different granularity. Further analysis in the characteristics of clusters is presented in next subsection.

\subsubsection{Further multimodal analysis}

Each transfer method is different in nature and can be exploited 
differently. 
Below, we analyze target transfer and structure transfer. 

\begin{figure}[tb]
\centering
a)
\epsfig{file=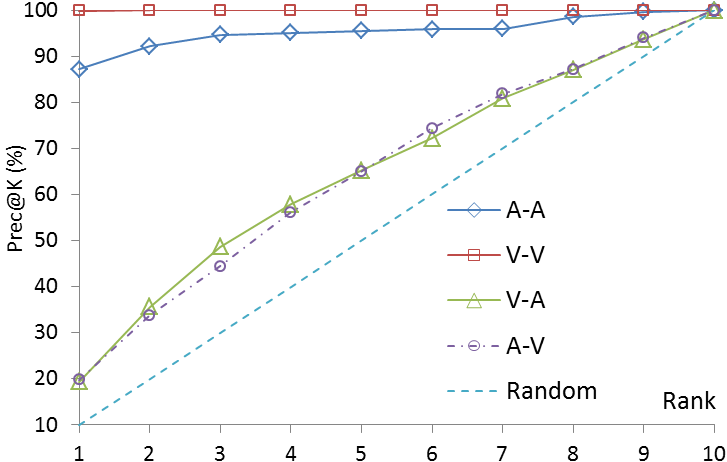,height=25mm}
b)
\epsfig{file=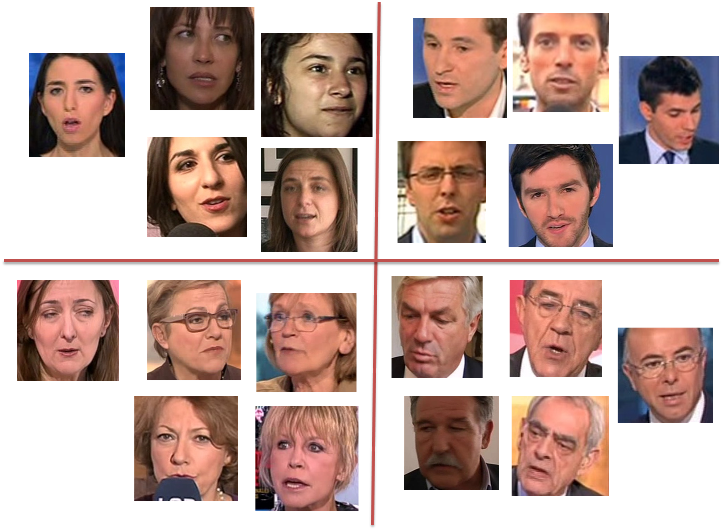,height=25mm}
\caption{Analysis of different transferring type. (a) Prec@K of cross modal id retrieval using target transfer, 
(b) visualization of shared identities in 4 clusters across both modalities.}
\label{fig:type_graph}
\end{figure}

\mypartitle{Cross modal retrieval.} One interesting potential of target embedding transfer is the ability to connect a voice to a face of the same identity. To explore this aspect, we formulate a retrieval experiment:
given 1 instance of the source embedding domain (voice or face), 
its distances to the embedding of 1 correct identities and 9 distractors in the enrolled domain are computed 
and ranked accordingly. 
There are 4 different settings depending on the within or cross domain retrieval: audio-audio, visual-visual, audio-visual, and visual-audio.
Fig.~\ref{fig:type_graph}-(a) shows the average precision of 980 different runs when choosing from the 
top 1 to 10 ranked results (Prec@K).
Although the cross modal retrieval settings cannot compete with their single modality counterparts, 
they perform better than random chance and show consistency between the face embedding and speaker turn embedding. 
This proves that the two modalities cannot be coupled as in coupled matching learning but can be used as a regularizer of one another.

\mypartitle{Shared clusters across modalities.} Fig.~\ref{fig:type_graph}-(b) visualizes 4 clusters which share the most common identities across the 2 modalities, when using the face embedding and the speaker embedding with structure transfer. 
One can observe 2 distinct characteristics among the clusters which are automatically captured: gender and age. 
It is noteworthy that these characteristics are discovered without any supervision.

\section{Conclusion}
\label{sec:conclusion}

Inspired by state-of-the-art machine learning techniques, we have proposed three different approaches to transfer knowledge from a source face embedding to a target speaker turn embedding. 
Each of our approaches explore different properties of the embedding spaces at different granularity. 
The results show that our methods  improved speaker turn embedding in the tasks of verification and clustering.
This is particularly significant in cases of short utterances, an important situation that can be found in many dialog cases, \textit{e.g.} TV series, debates, or in multi-party human-robot interactions where backchannels and short answers/utterances are very frequent.
The embedding spaces can also provide potential discovery of latent characteristics and a unified crossmodal combination. 
Another advantage of the transfer learning approaches is that each modality can be trained independently with their respective data,
thus allowing future extension using advance learning techniques or more available data. 

In the future, experiments with more complicated tasks such as person diarization or large scale indexing can be performed to explore the possibilities of each proposal. Also, working with other corpora in different languages is an interesting direction.

%
\bibliographystyle{abbrv}
\bibliography{main}  

\end{document}